\documentclass[letterpaper, 10 pt, conference]{ieeeconf}  

\IEEEoverridecommandlockouts                              
\usepackage[bookmarks=true]{hyperref}
\usepackage{tabularx}
\usepackage{graphicx}
\usepackage{epsfig}
\usepackage{subfig}
\usepackage{cite}
\usepackage{float}
\usepackage{ifthen}
\usepackage{xcolor}
\usepackage{ntheorem}
\theoremseparator{:}
\usepackage{soul}
\usepackage{colortbl}
\usepackage{booktabs}

\usepackage[font=footnotesize]{caption}
\usepackage[flushleft]{threeparttable}
\usepackage{enumerate}
\usepackage{algorithm}
\usepackage{algpseudocode}
\algrenewcommand\algorithmicindent{1.2em}
\usepackage{amsmath,amssymb}

\usepackage{censor}

\newif\ifanonymous
\anonymousfalse   % Use this for camera-ready

\ifanonymous
    
\else
    \StopCensoring
    
\fi

\newcommand{\fig}[1]{Fig.~\ref{#1}}

\def\ie{\emph{i.e., }} 
\title{\LARGE \bf Let the Body Follow: Coupled Egocentric Control for Whole-Body Robot Teleoperation}

\ifanonymous
\author{Anonymous Authors}
\else
\author{Tsung-Chi Lin$^{1}$, Yichen Xie$^{2}$, and Chien-Ming Huang$^{2}$%
\thanks{This work was supported by the Malone Center for Engineering in Healthcare at Johns Hopkins University.}%
\thanks{$^{1}$Department of Computer Science, New Jersey Institute of Technology, Newark, NJ, USA.
{\tt\small tsungchi.lin@njit.edu}}%
\thanks{$^{2}$Department of Computer Science, Johns Hopkins University, Baltimore, MD, USA.
{\tt\small \{yxie78, chienming.huang\}@jhu.edu}}%
}
\fi

\begin{document}

\bstctlcite{IEEEexample:BSTcontrol}

\maketitle
\thispagestyle{empty}
\pagestyle{empty}
% % % % % % % % % % % % % % % % % % % % % % % % % % % % % % % % % 

%%%=============================================
\begin{abstract}
%%%=============================================

Whole-body teleoperation requires users to coordinate perception, manipulation, posture, and mobility across multiple robot components. This coordination is difficult because users must simultaneously control the robot's head, arms, torso, and base while maintaining task awareness and avoiding kinematic or environmental constraints. In this paper, we propose \textit{coupled egocentric control}, a body-following teleoperation approach in which the robot's torso and base automatically respond to the operator's head and arm motions. Rather than requiring explicit touchpad commands for every torso or base adjustment, the system lets users focus on gaze and hand control: head pitch adjusts torso height, head yaw drives base rotation, end-effector height adjusts torso motion, and end-effector workspace boundaries trigger base translation. We evaluate this approach in a user study on whole-body teleoperation of a TIAGo mobile manipulator for home-care-inspired tasks. Compared with a baseline hybrid interface, coupled egocentric control improves object manipulation efficiency, reduces button-based control effort and arm singularities, lowers mental demand and overall workload, and increases ease of use, ease of learning, confidence, and user preference for torso and base control.

%%%=============================================
\end{abstract}
%%%=============================================

%%%=============================================
\section{Introduction}\label{sec:intro}
%%%=============================================

Mobile manipulators and humanoid robots are increasingly expected to operate in physically situated environments, from industrial settings~\cite{chappellet2023humanoid,tong2024advancements} to home care~\cite{andtfolk2022humanoid,lee2024my}. These robots can perceive, navigate, reach, manipulate, and reposition their bodies, making them promising platforms for remote assistance in complex real-world tasks. However, these same capabilities also make teleoperation difficult. To operate the whole body of a robot, users must coordinate multiple interdependent components---including the head, arms, torso, and mobile base---while maintaining situational awareness, avoiding obstacles, preventing awkward arm configurations, and completing the task efficiently. As a result, whole-body teleoperation is not simply a problem of providing access to more degrees of freedom; it is a coordination problem in which perception, manipulation, posture, and mobility must be controlled together without overwhelming the user.

Existing approaches to whole-body robot teleoperation typically rely on either \textit{free-form} control~\cite{koenemann2014real,terlemez2014master}, which maps the user's body or device motion to the robot and offers expressive control, or \textit{constrained} control~\cite{yamashita2016remote}, which restricts motion to predefined commands or axes for greater stability and precision. Free-form interfaces can be intuitive for controlling high-degree-of-freedom components such as the head and arms, but they can be difficult to manage and may require expensive or specialized hardware~\cite{dafarra2024icub3}. Constrained interfaces, in contrast, are often easier to stabilize for single-axis motion such as torso elevation or base translation, but they can limit the fluid use of the robot's full body. Hybrid interfaces combine these strengths by using free-form input for perception and manipulation while relying on constrained input for posture and mobility. However, even hybrid interfaces often require users to manually switch attention between control channels, such as moving the arms, adjusting the torso, rotating the base, and repositioning the robot. This manual coordination can increase cognitive workload, disrupt task flow, and reduce efficiency, especially in cluttered or multi-step manipulation tasks~\cite{lin2022intuitive}.

\begin{figure}[t]
    \centering
    \includegraphics[width=1\linewidth]{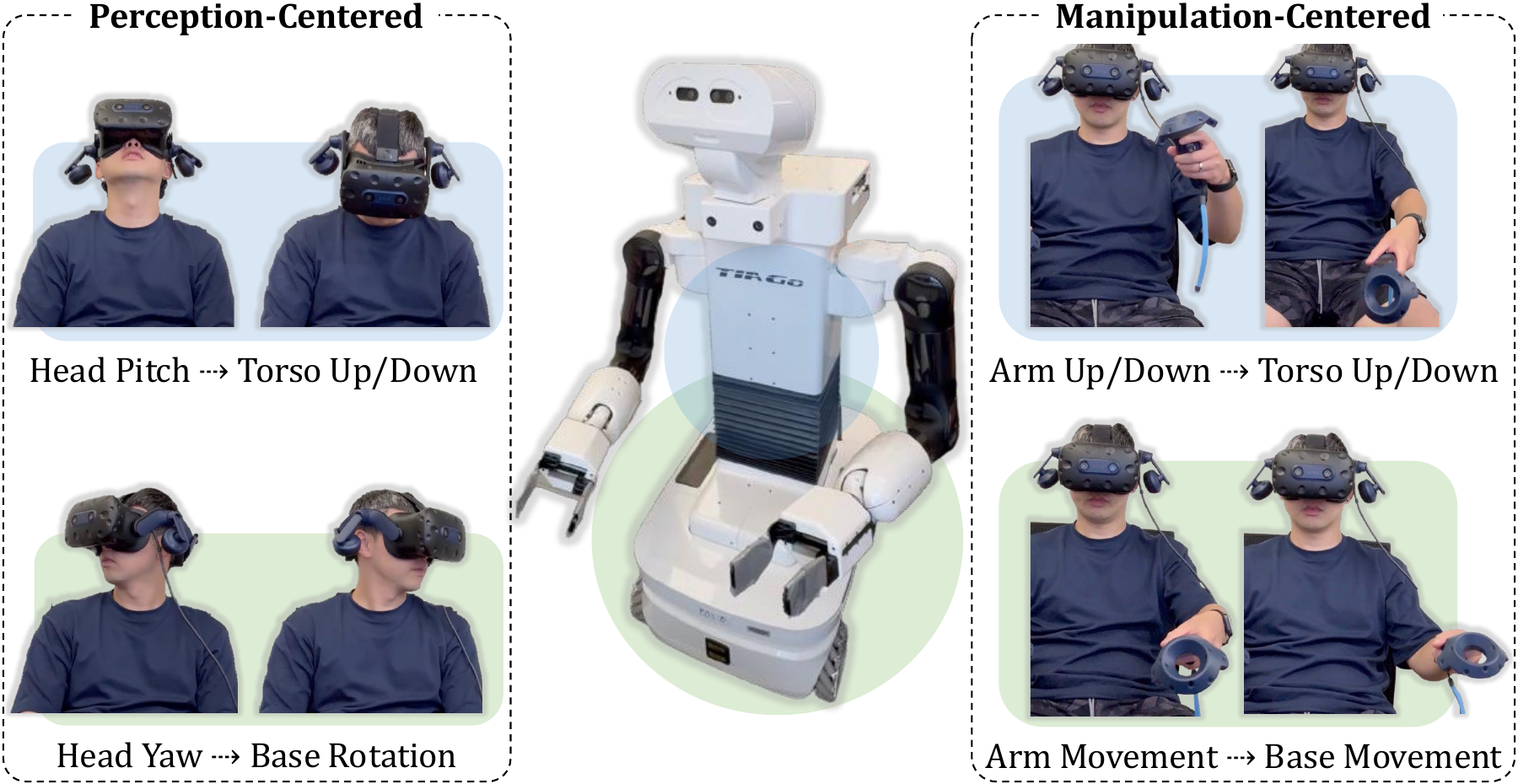}
    \caption{Our proposed coupled egocentric control system lets the robot body follow the operator's gaze and hands. Head motions support perception-centered control by adjusting torso height and base rotation, while arm motions support manipulation-centered control by adjusting torso height and base translation.}
    \label{fig:cover}
    \vspace{-3ex}
\end{figure} 

This paper introduces \textit{coupled egocentric control} for whole-body robot teleoperation (\fig{fig:cover}). The central idea is simple: let the body follow the operator's gaze and hands. Instead of asking users to explicitly command the torso and base whenever the viewpoint or manipulation workspace becomes insufficient, our system treats head and arm motion as signals of the operator's perceptual and manipulation intent, using them to automate supportive torso and base adjustments while preserving direct user control over the robot's gaze and hands. Specifically, our approach integrates two forms of body-following control: in \textit{perception-centered control}, the torso moves up or down when the head tilts beyond predefined thresholds to extend the vertical viewpoint, and the base rotates when the head pans beyond left or right thresholds to extend the field of view; in \textit{manipulation-centered control}, the torso moves up or down when an end effector reaches vertical workspace boundaries, and the base moves forward, backward, or sideways when an end effector reaches translational workspace boundaries. By coupling these motions, the system reduces the need for explicit torso/base commands, helps avoid arm overstretching and singularities, and supports continuous transitions between looking, reaching, and repositioning.

We evaluate coupled egocentric control on a TIAGo mobile manipulator in home-care-inspired tasks, comparing it with a baseline hybrid VR interface in which the robot's head and dual arms are controlled through an HTC Vive Pro 2 headset and handheld controllers, while the torso and base are controlled through touchpads. The study consisted of two phases: Phase I compared the baseline and coupled egocentric interfaces across three tasks emphasizing base movement, torso adjustment, and base rotation; Phase II examined user preference in a more complex cleaning task involving multiple objects and workspaces. Results show that coupled egocentric control improves object manipulation efficiency, reduces button-based control effort and arm singularities, lowers mental demand and overall workload, and increases ease of use, ease of learning, confidence, and preference for torso adjustment and base rotation.

The main contributions of this paper are:

\noindent(C1) \textit{Coupled egocentric control} for allowing the robot body to follow the operator's gaze and hands through head--torso/base and arm--torso/base coupling;

\noindent(C2) \textit{VR-based whole-body teleoperation} for implementing this control framework with real-time visual feedback, obstacle awareness, and collision-avoidance support; and

\noindent(C3) \textit{User study evaluation} for demonstrating improved manipulation efficiency, reduced control effort and workload, and stronger user preference in home-care-inspired whole-body teleoperation tasks.

%%%=============================================
\section{Background and Related Work}\label{sec:related}
%%%=============================================

\subsection{Whole-Body Teleoperation Interfaces}

Whole-body teleoperation requires users to coordinate multiple robot components, including perception through the head or camera, manipulation through the arms, posture through the torso, and mobility through the base or legs. Existing interfaces generally fall into constrained, free-form, and hybrid control approaches. \textit{Constrained control} simplifies operation by limiting the robot's degrees of freedom or restricting motion to predefined commands, supporting stable control for tasks such as manipulation~\cite{mavridis2015subjective} or navigation~\cite{huang2020telelocomotion}. However, this simplicity can limit the robot's full capabilities in tasks that require coordinated looking, reaching, repositioning, and interaction.

In contrast, \textit{free-form control} allows users to command robot motion through body, hand, or device movements, using wearable sensors~\cite{zhou2019iot}, exosuits~\cite{dafarra2024icub3}, or vision-based tracking~\cite{rakita2017motion}. These interfaces can be intuitive for high-DoF components such as the head and arms, but they also require users to manage many control dimensions simultaneously, which can increase cognitive and physical effort. \textit{Hybrid control} approaches combine the strengths of both paradigms, for example using VR pose tracking for head and arm control while relying on touchpads or discrete inputs for torso and base motion in mobile manipulators~\cite{bejczy2020mixed} and humanoid robots~\cite{wonsick2021human}. Yet, even hybrid interfaces often leave the coordination problem to the user: operators must manually switch attention among gaze, arm motion, torso adjustment, and base movement, which can interrupt task flow and increase workload.

\subsection{Semi-Autonomous Loco-Manipulation}

To reduce this coordination burden, prior work has explored semi-autonomous support for loco-manipulation, where locomotion and manipulation are coordinated to maintain reachability, visibility, and task progress. Manipulation-centered approaches allow the robot to coordinate reaching and walking around the user's manipulation input~\cite{fukumoto2004hand}, while multi-centered approaches incorporate both arm motion and broader human-body motion for integrated navigation and manipulation~\cite{ha2015whole,wu2019teleoperation}. Other systems provide higher-level autonomy for task-specific whole-body motion~\cite{stilman2008humanoid} or optimize robot motion around object interaction and transportation~\cite{murooka2021humanoid}. Although these approaches demonstrate the value of shared autonomy, many are designed around specific tasks, objects, or predefined coordination policies. Less attention has been paid to how the operator's natural egocentric inputs---especially head motion and hand motion---can continuously guide supportive body movements. In whole-body teleoperation, head motion often indicates perceptual intent, such as looking higher, lower, left, or right, while arm motion often indicates manipulation intent, such as extending reach or approaching a target. This paper builds on this observation by introducing \textit{coupled egocentric control}, a body-following approach that couples head motion with torso elevation and base rotation, and end-effector motion with torso adjustment and base translation. This design allows users to focus on where the robot looks and reaches while the rest of the robot body follows to maintain viewpoint, reachability, and maneuverability.

%%%=============================================
\section{Whole-Body Robot Teleoperation}\label{sec:proposed}
%%%=============================================

We present a whole-body teleoperation system for a TIAGo OMNI++ mobile manipulator. The system builds on a hybrid VR interface that provides direct control of the robot's head and arms while using constrained inputs for the torso and base. We first summarize this baseline framework and its visual/safety support, then introduce our proposed \textit{coupled egocentric control}, in which the robot body follows the operator's gaze and hands by automatically coordinating torso and base motion with head and arm movements.

%%% --------------------------------------------
\subsection{Hybrid Control Framework}
%%% --------------------------------------------

\begin{figure}[b]
    \centering
    % \vspace{-1ex}
    \includegraphics[width=1\linewidth]{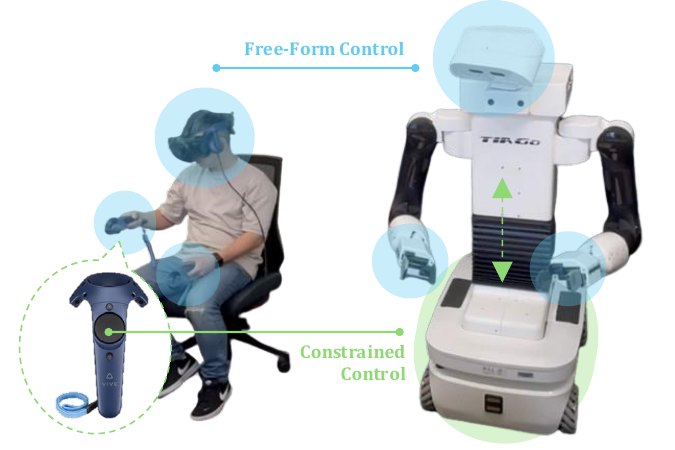}
    \caption{The baseline hybrid teleoperation interface captures user head and hand movements to control the robot's head and arms, while touchpads on the handheld controllers control the robot's torso and base.}
    \label{fig:interface}
    % \vspace{-2ex}
\end{figure}

The baseline interface uses a hybrid control framework that combines free-form pose control with constrained touchpad control (\fig{fig:interface}). The robot's head and dual arms are controlled through the pose of an HTC Vive Pro 2 headset and two handheld controllers, while the torso and mobile base are controlled through the touchpads. This design provides intuitive control for high-DoF perception and manipulation, while preserving stable single-axis commands for torso elevation and base motion.

For head control, the user's headset rotation is mapped to the robot's 2-DoF pan--tilt head. A $30^\circ$ downward pitch offset is applied so that the robot's default gaze aligns with the manipulation workspace, reducing the need for users to maintain a bent-neck posture during prolonged operation. For arm control, each handheld controller specifies the desired Cartesian pose of the corresponding end effector, and TRAC-IK~\cite{beeson2015trac} computes the robot arm configuration. Users can pause or activate arm control with the grip button, reset the arm to a home configuration with the menu button, and open or close the gripper with the trigger button.

For torso and base control, touchpads provide constrained commands. The right touchpad moves the torso up or down in $0.05$~m increments, with limits that prevent collision with the base. The left touchpad controls base translation, including forward, backward, lateral, and diagonal motions, while left/right presses on the right touchpad control base rotation. Algorithm~\ref{alg:hybrid} summarizes the baseline control loop.

%%% --------------------------------------------
\subsection{Visual Feedback and Safety Awareness}
%%% --------------------------------------------

\begin{figure}[t]
    \centering
    % \vspace{-1ex}
    \includegraphics[width=1\linewidth]{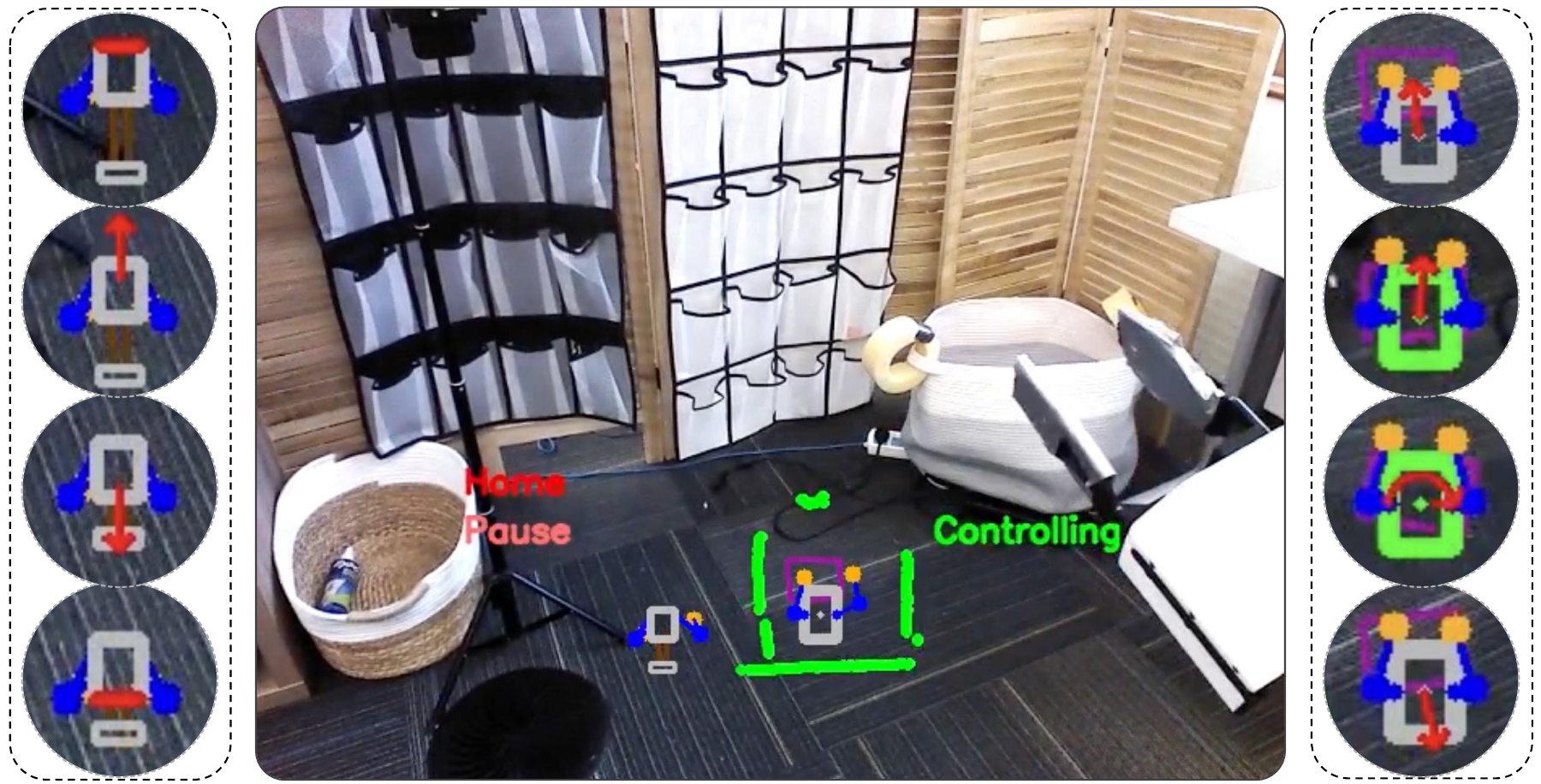}
    \caption{The advanced GUI integrates a video stream, arm control states, and both back and top views of the robot's mini model, displaying the robot's arm pose, torso height, and surrounding obstacles. Red arrows indicate control inputs, with red lines for the torso’s vertical limits. The base is colored green or gray to signify ``fast'' or ``slow'' mode depending on whether the center of the base is within or outside the purple area, respectively.}
    \label{fig:gui}
    \vspace{-3ex}
\end{figure}

A graphical user interface (GUI) integrated real-time video feedback from an RGB camera built into the TIAGo's head, overlaying the robot's operational states and streaming them to the HTC Vive headset via the User Datagram Protocol. In addition to displaying the arm control states (\ie showing control, pause, and home statuses), we advanced the display of the robot's states by providing back and top views of the robot's mini model, which monitored its arm pose, torso height, base speed, and surrounding obstacles (\fig{fig:gui}). 

\noindent
\textbf{Visualization of Arm Pose and Torso Elevation:}
Simplified robot arms with fixed shoulder joints (small blue dots) and real-time elbow (large blue circles) and wrist (large orange circles) joints were added to both the back and top views of the mini robot model to enhance arm pose awareness; monitoring arm pose is crucial in preventing overstretching and awkward configurations during robot operation. Torso height was indicated by the height of the upper part of the robot in the back view, with red arrows showing control inputs and red lines marking the upper and lower boundaries to indicate the limits of adjustable range.

\noindent
\textbf{Visualization of Base Control:}
The top view of the robot's mini model displayed the base control inputs, with red arrows indicating translational movement directions and red curved arrows showing rotational directions. We implemented dynamic scaling of the base speed based on the workspace area. Specifically, when the robot base was near the center of the workspace---an uncluttered, obstacle-free area (purple region)---the base operated in fast mode (colored green) for more efficient navigation; however, when the robot base approached any surroundings, indicating a cluttered area, the base significantly reduced its speed (turning gray) to prevent potential collisions and enhance control precision.

\noindent
\textbf{Visualization of Proximal Obstacles:}
To provide real-time feedback on the robot's movements and obstacle awareness, we created a dynamic mini-map in the top view that continuously updated to show obstacles within the robot's workspace (\fig{fig:gui}), with a red color change indicating that the robot was too close to a particular obstacle. In addition to displaying the distance to obstacles, we further provided the precise height of nearby tables (yellow lines) and a shelf (yellow shape) in the back view when the robot was near them. This detailed information, combined with the robot's mini model, allowed for a clearer understanding of the relationship between the robot's arms and surrounding objects, significantly improving collision avoidance and object interaction.

\begin{algorithm}[t]
\caption{Baseline Hybrid Whole-Body Teleoperation}
\label{alg:hybrid}
\begin{algorithmic}[1]
\State Initialize VR devices, robot state, and GUI
\While{teleoperation is active}
    \State Read headset orientation $R_h$
    \State Read controller poses $\mathbf{x}^{L}_{c}, \mathbf{x}^{R}_{c}$
    \State Command head pan--tilt from $R_h$ with pitch offset
    \For{$i \in \{L,R\}$}
        \If{arm control $i$ is active}
            \State Map $\mathbf{x}^{i}_{c}$ to desired pose $\mathbf{x}^{i}_{ee}$
            \State $\mathbf{q}^{i}_{a} \gets \mathrm{IK}(\mathbf{x}^{i}_{ee})$
            \State Send $\mathbf{q}^{i}_{a}$ and gripper command
        \EndIf
    \EndFor
    \State $\dot{T}_{z} \gets$ right-touchpad up/down command
    \State $\mathbf{v}_{b} \gets$ left-touchpad translation command
    \State $\dot{\psi}_{b} \gets$ right-touchpad rotation command
    \State Apply scaling, damping, and safety constraints
    \State Send torso/base commands and update GUI
\EndWhile
\end{algorithmic}
\end{algorithm}

\begin{figure}[b]
    \centering
    % \vspace{-1ex}
    \includegraphics[width=1\linewidth]{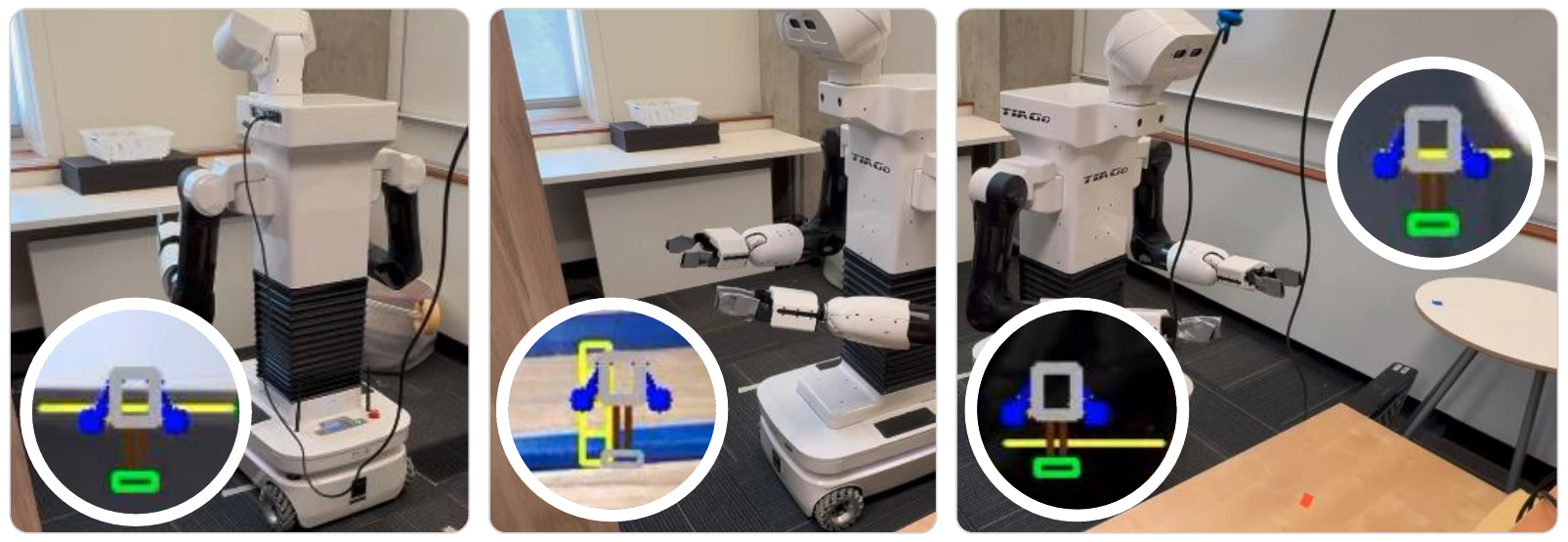}
    \caption{The GUI displays the heights of nearby tables (yellow lines) and the shelf (yellow shape) in the back view when the robot is close to them, enhancing collision avoidance and object interaction.}
    \label{fig:obstacles}
    % \vspace{-2ex}
\end{figure} 

%%% --------------------------------------------
\subsection{Collision Avoidance and Motion Damping}
%%% --------------------------------------------

\begin{figure}[t]
    \centering
    % \vspace{-1ex}
    \includegraphics[width=1\linewidth]{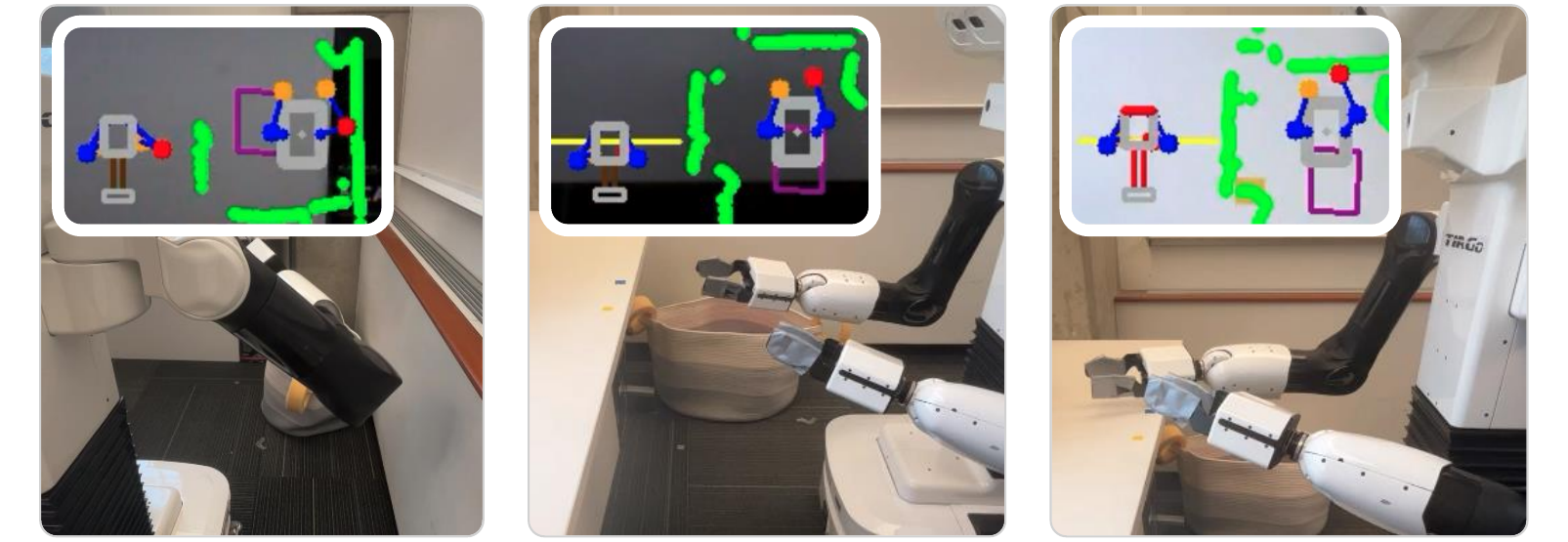}
    \caption{The arm collision avoidance system monitors the robot's elbows and end effectors. The GUI provides alerts while the system takes action (\ie pausing arm control or deactivating certain motions) to prevent collisions.}
    \label{fig:collision}
    \vspace{-3ex}
\end{figure} 

In addition to an alert indicating a nearby obstacle (\ie the obstacles in the top view), we implemented a damping feature: When the robot's base approached an obstacle, the movement in that direction was gradually reduced to zero, preventing further motion toward the obstacle while allowing movement in all other directions.
During object manipulation, the system monitored the robot's elbows and end effectors to ensure they did not come too close to obstacles (\ie walls, tables, or shelves) and took action to prevent collisions (\fig{fig:collision}); specifically, when an elbow was near an obstacle, the corresponding elbow joint in both the back and top views of the GUI turned red, and arm control was paused. For end effector collision avoidance, if the arm movement was insufficient to avoid an obstacle (\ie the end effector was lower than a table or shelf, or above a table but moving downward), motion in the direction toward the obstacle was reduced to zero, preventing further movement toward it while allowing motion in all other directions; related movements that would result in the same collision risk were also deactivated (\ie the base could not move forward when the end effector was lower than the table, or the torso could not lower itself when the end effector was above and close to the table) and the GUI alerted the user to the risk by turning the robot's wrist joints and torso links red in both the back and top views.

%%% --------------------------------------------
\subsection{Coupled Egocentric Control}
%%% --------------------------------------------

Although the hybrid interface provides direct access to the robot's whole body, it still requires users to manually coordinate head, arms, torso, and base. This coordination can interrupt the flow of teleoperation: users may need to stop reaching to adjust torso height, stop looking to rotate the base, or issue repeated touchpad commands to keep the robot close to the workspace. To reduce this coordination burden, we propose \textit{coupled egocentric control}: a body-following strategy in which head and arm motions remain under direct user control, while the torso and base automatically follow these motions to support perception and manipulation.

Let $\theta_t$ and $\theta_p$ denote the robot head tilt and pan angles after headset-to-head mapping, and let $\mathbf{p}^{i}_{ee}=[x^{i}_{ee},y^{i}_{ee},z^{i}_{ee}]^\top$ denote the position of end effector $i \in \{L,R\}$ in the robot base frame. The controller generates three supportive body commands: torso velocity $\dot{T}_{z}$, base translational velocity $\mathbf{v}_{b}=[v_x,v_y]^\top$, and base rotational velocity $\dot{\psi}_{b}$. These commands are produced by two coupled policies: perception-centered coupling driven by head motion and manipulation-centered coupling driven by end-effector motion.

We define a threshold function
\begin{equation}
\Gamma(s;\tau,u)=
\begin{cases}
u, & s>\tau,\\
-u, & s<-\tau,\\
0, & \mathrm{otherwise},
\end{cases}
\label{eq:threshold}
\end{equation}
where $s$ is the input signal, $\tau$ is the activation threshold, and $u$ is the output velocity.

\noindent
\textbf{Perception-Centered Coupling:}
Head motion provides an egocentric signal of where the operator wants to look. When the head tilts near its upper or lower range, the torso moves to extend the vertical viewpoint; when the head pans near its left or right range, the base rotates to extend the horizontal field of view. The head-driven torso and base rotation commands are:
\begin{equation}
\dot{T}^{h}_{z}
=
\Gamma(\theta_t;\lambda_\theta\theta^{\max}_{t},v_T),
\label{eq:head_torso}
\end{equation}
\begin{equation}
\dot{\psi}_{b}
=
\Gamma(\theta_p;\lambda_\theta\theta^{\max}_{p},\alpha_M\omega_b),
\label{eq:head_base}
\end{equation}
where $v_T$ is the torso speed, $\omega_b$ is the maximum base rotational speed, $\lambda_\theta=0.5$ is the activation threshold ratio, and $\alpha_M$ is a speed-scaling factor determined by the current base mode (\ie fast in open space and slow near obstacles).

\begin{figure}[b]
    \centering
    \includegraphics[width=1\linewidth]{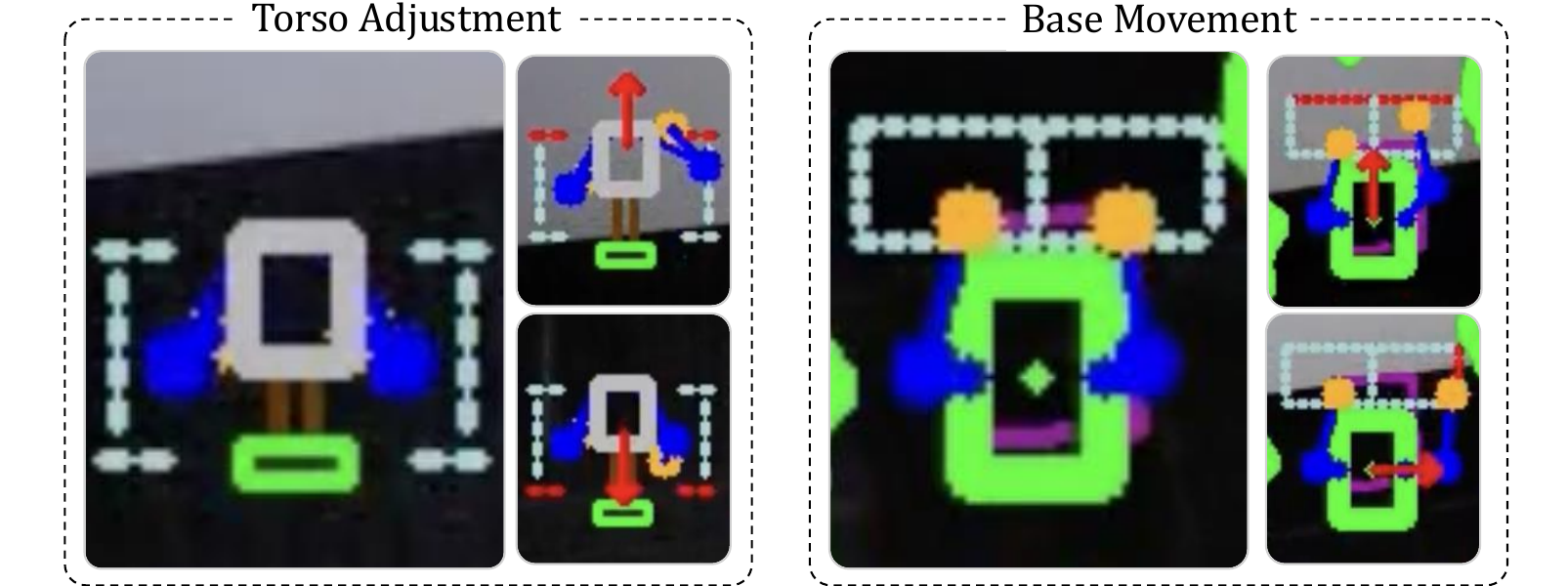}
    \caption{Manipulation-centered coupling in coupled egocentric control. Torso adjustment and base translation are triggered when the robot's end effector reaches predefined workspace boundaries.}
    \label{fig:egocentric}
\end{figure}

\noindent
\textbf{Manipulation-Centered Coupling:}
End-effector motion provides an egocentric signal of where the operator wants the robot to reach. When an end effector approaches the vertical boundary of the manipulation workspace, the torso moves to extend reach; when it approaches the horizontal boundary, the base translates to keep the arm within a maneuverable region. For each active arm $i$, the arm-driven torso command is:
\begin{equation}
\dot{T}^{a,i}_{z}=
\begin{cases}
+v_T, & z^{i}_{ee}>z^{\max}_{ee},\\
-v_T, & z^{i}_{ee}<z^{\min}_{ee},\\
0, & \mathrm{otherwise}.
\end{cases}
\label{eq:arm_torso}
\end{equation}
The arm-driven base translation command is:
\begin{equation}
\small
\mathbf{v}^{i}_{b}=
\begin{cases}
[+v_x,0]^\top, & x^{i}_{ee}>x^{\max}_{ee},\\
[-v_x,0]^\top, & x^{i}_{ee}<x^{\min}_{ee},\\
[0,+v_y]^\top, & y^{i}_{ee}>y^{\max}_{ee},\\
[0,-v_y]^\top, & y^{i}_{ee}<y^{\min}_{ee},\\
[0,0]^\top, & \mathrm{otherwise}.
\end{cases}
\label{eq:arm_base}
\end{equation}
where $[x^{\min}_{ee},x^{\max}_{ee}]$, $[y^{\min}_{ee},y^{\max}_{ee}]$, and $[z^{\min}_{ee},z^{\max}_{ee}]$ define the manipulation-centered control region. These boundaries are displayed as light blue dotted lines in the GUI (\fig{fig:egocentric}). In our implementation, positive and negative values of $v_x$ and $v_y$ correspond to the robot's forward/backward and left/right base motions according to the TIAGo base-frame convention.

The final torso command combines perception-centered and manipulation-centered coupling with saturation:
\begin{equation}
\dot{T}_{z}
=
\mathrm{sat}_{[-v_T,v_T]}
\left(
\dot{T}^{h}_{z}
+
\sum_{i\in\mathcal{A}}\dot{T}^{a,i}_{z}
\right),
\label{eq:torso_final}
\end{equation}
where $\mathcal{A}$ is the set of active arms. For base translation, the robot moves in only one direction at a time to maintain predictable behavior. If multiple boundaries are activated, the command is selected using the priority order backward, sideways, and forward. This priority helps the robot first move away from overextended arm configurations before making lateral or forward adjustments.

The final base translation command is selected from the active arm-driven candidates:
\begin{equation}
\mathbf{v}_{b}=
\Pi_{\rho}
\left(
\left\{\mathbf{v}^{i}_{b}\mid i\in\mathcal{A},\ \mathbf{v}^{i}_{b}\neq \mathbf{0}\right\}
\right),
\label{eq:base_final}
\end{equation}
where $\Pi_{\rho}(\cdot)$ selects one nonzero command according to the priority order $\rho$: backward, sideways, and forward.

Together, these coupling rules allow the user to continue controlling the robot's gaze and hands while the robot body follows to preserve viewpoint, reachability, and maneuverability. Compared with the baseline hybrid interface, coupled egocentric control reduces the need for explicit touchpad commands and supports smoother transitions between looking, reaching, and repositioning.

To summarize the difference between the baseline hybrid interface and our coupled egocentric controller, Table~\ref{tab:control_mapping} compares the control mappings for each robot component. The key distinction is that the baseline requires explicit touchpad commands for torso and base motion, whereas coupled egocentric control allows these supporting body motions to follow the operator's gaze and hands.

\begin{table}[t]
\centering
\caption{Baseline and coupled egocentric control mappings.}
\label{tab:control_mapping}
\footnotesize
\begin{tabularx}{\columnwidth}{@{}p{0.24\columnwidth}p{0.24\columnwidth}X@{}}
\toprule
\textbf{Component} & \textbf{Baseline} & \textbf{Coupled Egocentric Control} \\
\midrule
Head 
& Headset pose 
& Direct control; head pitch/yaw also drives torso height and base rotation. \\

Arms 
& Controller pose 
& Direct control; end-effector motion also drives torso height and base translation. \\

Torso 
& Right touchpad 
& Follows head pitch or end-effector height. \\

Base rotation 
& Right touchpad 
& Follows head yaw. \\

Base translation 
& Left touchpad 
& Follows end-effector workspace boundaries. \\
\bottomrule
\end{tabularx}
\vspace{-2ex}
\end{table}

%%%=============================================
\section{User Study}\label{sec:exp}
%%%=============================================

We conducted a human participant study to evaluate whether coupled egocentric control improves the efficiency, usability, and perceived workload of whole-body robot teleoperation. The study compared our proposed interface with the baseline hybrid interface and examined both task performance and user preference in home-care-inspired manipulation scenarios.

\noindent
\textbf{Participants:}
We recruited 12 participants (5 male, 7 female) from a local university campus, aged 19 to 39 (\textit{M} = 27.92, \textit{SD} = 5.74). All participants had previously completed a study using the same baseline hybrid interface for coordinated whole-body teleoperation. In that prior study, they underwent systematic, curriculum-based training covering basic manipulation skills (\ie unimanual and bimanual manipulation of solid and deformable objects) and coordinated whole-body control (\ie head--arm, torso--arm, base--arm, and head--torso--base--arm coordination). Participants therefore had high familiarity with the baseline interface (\textit{M} = 4.33, \textit{SD} = 0.75) on a five-point scale, where 5 indicated high familiarity. This prior experience provided a conservative comparison for evaluating whether coupled egocentric control offers benefits beyond a well-practiced manual baseline. The study lasted 90 minutes, and each participant received \$15 for their time.

\begin{figure}[b]
    \centering
    \includegraphics[width=1\linewidth]{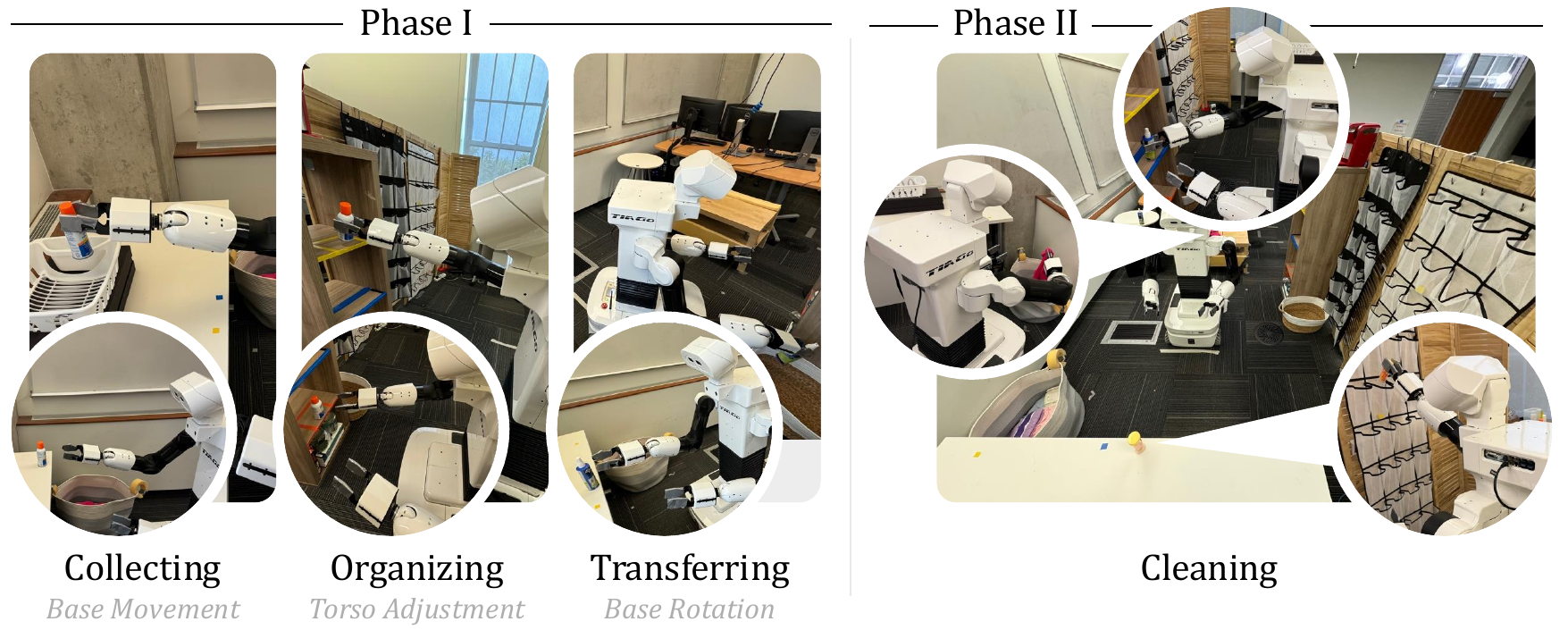}
    \caption{Home-care-inspired tasks used in the study. Phase I included isolated control tasks emphasizing base translation, torso adjustment, and base rotation. Phase II used a composite cleaning task to examine interface utilization and preference in a more realistic scenario.}
    \label{fig:tasks}
\end{figure}

\noindent
\textbf{Home-Care-Inspired Tasks:}
We designed a set of structured evaluation tasks to isolate specific whole-body control demands in \textit{Phase I} and a composite task to assess interface preference in a realistic home-care-inspired scenario in \textit{Phase II} (\fig{fig:tasks}). All object locations and target regions were fixed across participants.

In \textit{Phase I}, participants performed three short tasks using both the baseline and coupled egocentric interfaces. Each task was designed to emphasize a dominant control demand while minimizing the need for other body motions. First, in the \textbf{collecting task} (\textit{base-dominant}), a bottle was placed on the far-left side of a table. Participants moved the robot base laterally to reach the object and placed it into a bin located at the front center of the workspace. Torso motion was optional but not required, making base translation the primary control demand. Second, in the \textbf{organizing task} (\textit{torso-dominant}), participants moved a bottle from a lower shelf to an upper shelf directly above it, with minimal base displacement. This task primarily required torso extension and vertical reach adjustment, isolating torso control precision. Third, in the \textbf{transferring task} (\textit{rotation-dominant}), participants picked up a spray bottle from a table, rotated the robot approximately $90^\circ$ in place, and placed the object into a basket located in an adjacent workspace. This task emphasized in-place base rotation.

In \textit{Phase II}, participants performed a \textbf{cleaning task} as a composite whole-body teleoperation scenario. The task involved heterogeneous objects, including rigid bottles and a deformable cloth, and multiple workspaces, including tables of different heights, a multi-level shelf, and a wall-mounted organizer. Participants sequentially collected objects, relocated them across workspaces, and organized them into designated target locations. During this task, participants were allowed to switch freely between the baseline and coupled egocentric interfaces at any time. Task completion was defined as placing all objects in their corresponding target locations.

\noindent
\textbf{Experimental Procedure:}
After providing informed consent, participants received an explanation of both the baseline hybrid interface and the coupled egocentric interface. They then completed up to 20 minutes of hands-on practice with a training task that required moving a bottle between two tables of different heights. After training, participants completed two phases. In Phase I, they performed the three simple tasks with both interfaces, resulting in six trials in total (2 interfaces $\times$ 3 tasks); both task order and interface order were randomized. In Phase II, participants completed the complex cleaning task with no restrictions on sub-task order or arm usage, and could choose between the baseline and coupled egocentric control modes for torso and base operation. After each phase, participants completed the NASA-TLX and a questionnaire about usability and preference.

\noindent
\textbf{Measures and Analyses:}
In Phase I, we measured task completion time and object manipulation time to evaluate efficiency. Object manipulation time was defined as the duration from when the end effector entered a 0.3~m region around the target to the completion of the corresponding grasping or placing action. To evaluate control effort, we measured the number of controller button presses and the frequency with which the robot arms approached joint limits or singular configurations. In Phase II, we recorded task completion time and the duration of coupled egocentric control use. Subjective measures included NASA-TLX workload, ease of use, ease of learning, confidence, and control preference. We analyzed variance using an F-test and selected either Student's $t$-test or Welch's $t$-test depending on variance equality. We also used a mixed regression model to examine the relationship between coupled egocentric control use and both cleaning task time and NASA-TLX scores. Overall NASA-TLX score was computed using weighted subscales: mental demand = 5, physical demand = 1, temporal demand = 0, performance = 3, effort = 4, and frustration = 2.

%%%=============================================
\section{Results, Discussion, and Implications}\label{sec:res}
%%%=============================================

% % % % % % % - - - - - - - % % % % % % %
\subsection{Phase I: Baseline vs. Coupled Egocentric Control}
% % % % % % % - - - - - - - % % % % % % %

\begin{figure}[t]
    \centering
    \includegraphics[width=1\linewidth]{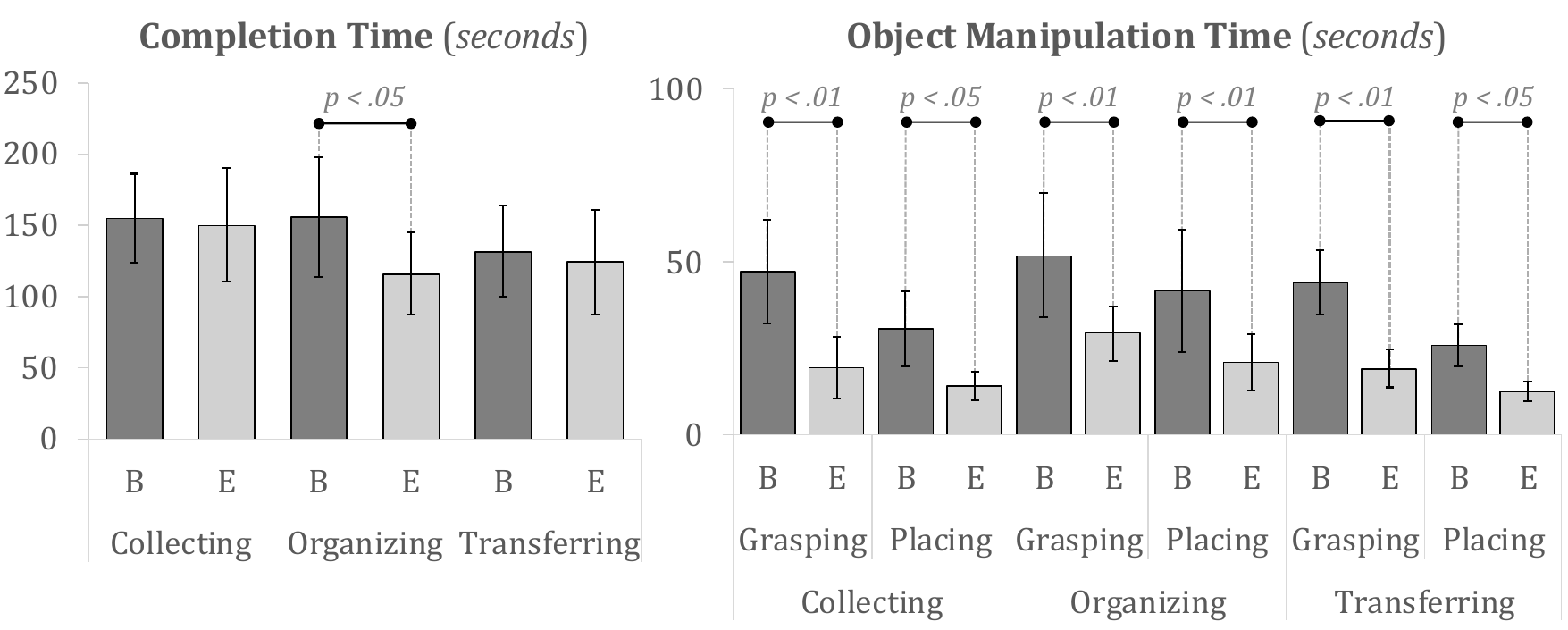}
    \caption{Comparison of task time and object manipulation time across the three home-care-inspired tasks between the baseline (B) and coupled egocentric (E) interfaces.}
    \vspace{-3ex}
    \label{fig:phase1-time}
\end{figure} 

\begin{figure}[b]
    \centering
    \includegraphics[width=1\linewidth]{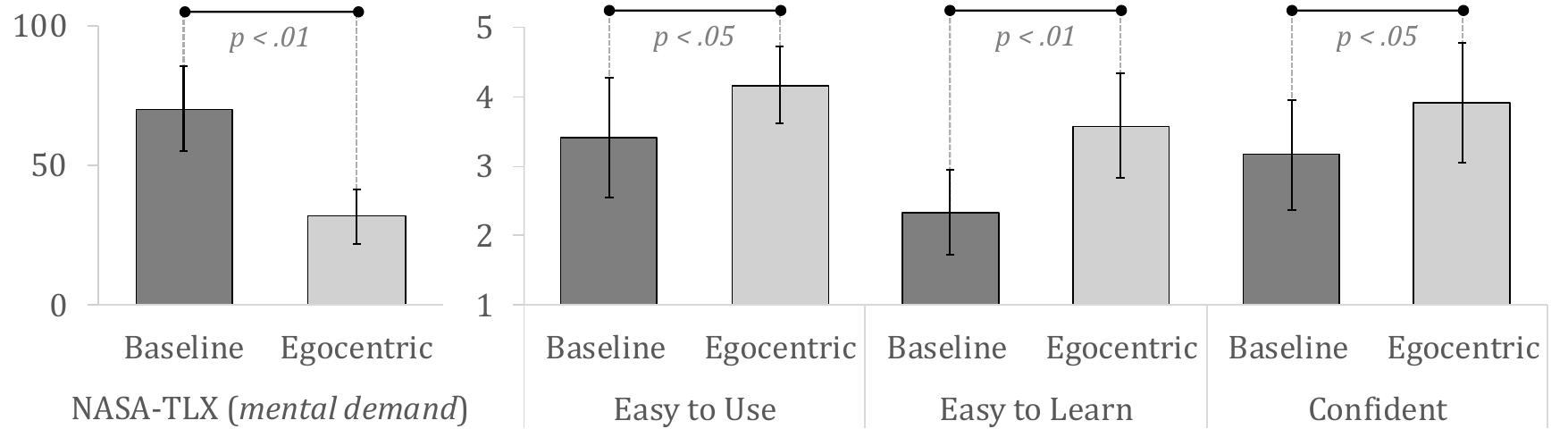}
    \caption{Subjective feedback on mental demand, ease of use, ease of learning, and confidence for the baseline and coupled egocentric interfaces.}
    \label{fig:phase1-subjective}
\end{figure} 

As shown in \fig{fig:phase1-time}, left, coupled egocentric control resulted in comparable task completion times for the collecting and transferring tasks, with no significant difference from the baseline, while significantly reducing completion time for the organizing task ($p < .05$). This suggests that body-following control was particularly beneficial when the task required frequent torso adjustment. More broadly, \fig{fig:phase1-time}, right shows that coupled egocentric control significantly reduced object manipulation time for both grasping and placing actions across all three tasks, indicating that automatically coordinating the torso and base with the operator's head and hand motions supported more efficient interaction with objects.

Coupled egocentric control also reduced control effort. Across all Phase I tasks, the baseline interface required \textit{five} times more button usage than the coupled egocentric interface. In addition, the robot arms encountered joint limits or singularities \textit{three} times more frequently with the baseline than with coupled egocentric control, suggesting that body-following adjustments helped maintain more maneuverable arm configurations. Subjective results further support these findings: coupled egocentric control significantly lowered participants' mental demand ($p < .01$) and increased ease of use ($p < .05$), ease of learning ($p < .01$), and confidence ($p < .05$) compared with the baseline interface (\fig{fig:phase1-subjective}).

\textit{Discussion ---} These results show that coupled egocentric control improves whole-body teleoperation by allowing the robot body to follow the operator's gaze and hands. Although total task completion time did not significantly improve for the collecting and transferring tasks, which involved substantial navigation and repositioning, object manipulation became faster across all tasks. This indicates that the proposed coupling was most effective during the interaction-rich portions of the tasks, where users needed to coordinate reaching, torso height, and base position. The reduction in button presses and arm singularities further suggests that the system offloaded low-level torso/base coordination while preserving direct user control over gaze and hands. Participants' comments reflected this benefit: \textit{``The coupled egocentric control helped me approach and grasp/place the target seamlessly at the same time''} and \textit{``It was useful that the base/torso moved in sync with the robot hand, as I often forgot which button to press to control the base/torso.''}

\textit{Design Implications ---}
The Phase I results suggest that coupled egocentric control is most beneficial during interaction-rich moments, when users must simultaneously manage reaching, torso height, and base position. Even when overall task completion time was comparable for navigation-heavy tasks, object manipulation time improved across all tasks, indicating that body-following support can reduce the coordination overhead around grasping and placing. This suggests that whole-body teleoperation interfaces should not only optimize navigation or manipulation separately, but should provide lightweight coupling mechanisms that preserve arm maneuverability and reduce the need for explicit posture and base commands during object interaction.

% % % % % % % - - - - - - - % % % % % % %
\subsection{Phase II: Utilization and Preference in a Complex Task}
% % % % % % % - - - - - - - % % % % % % %

\begin{figure}[b]
    \centering
    \includegraphics[width=1\linewidth]{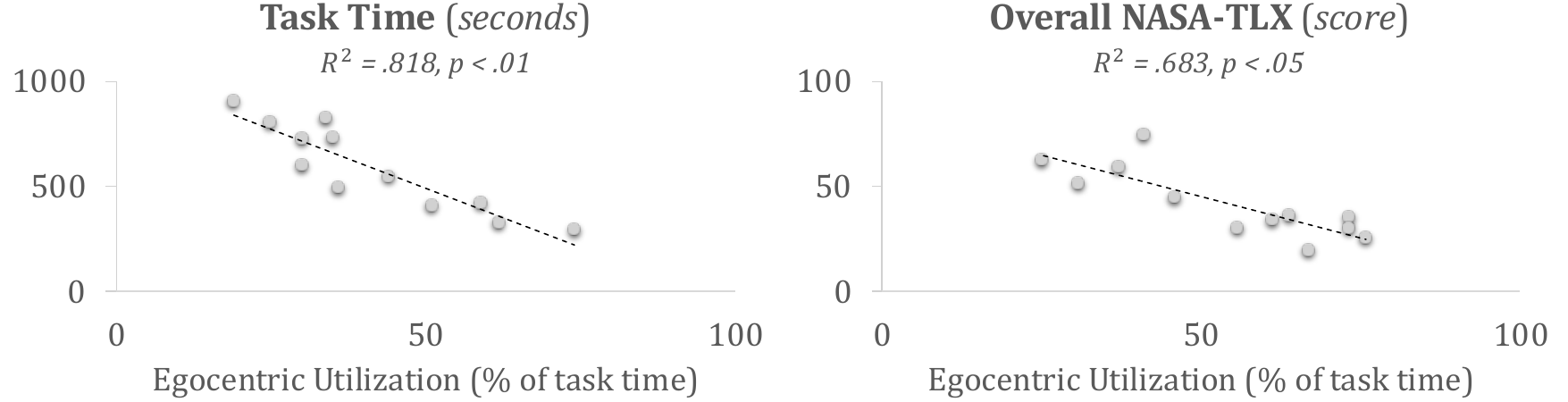}
    \caption{Correlation between coupled egocentric control utilization and both cleaning task time and overall NASA-TLX score.}
    \label{fig:phase2-regression}
\end{figure} 

\begin{figure}[t]
    \centering
    \includegraphics[width=1\linewidth]{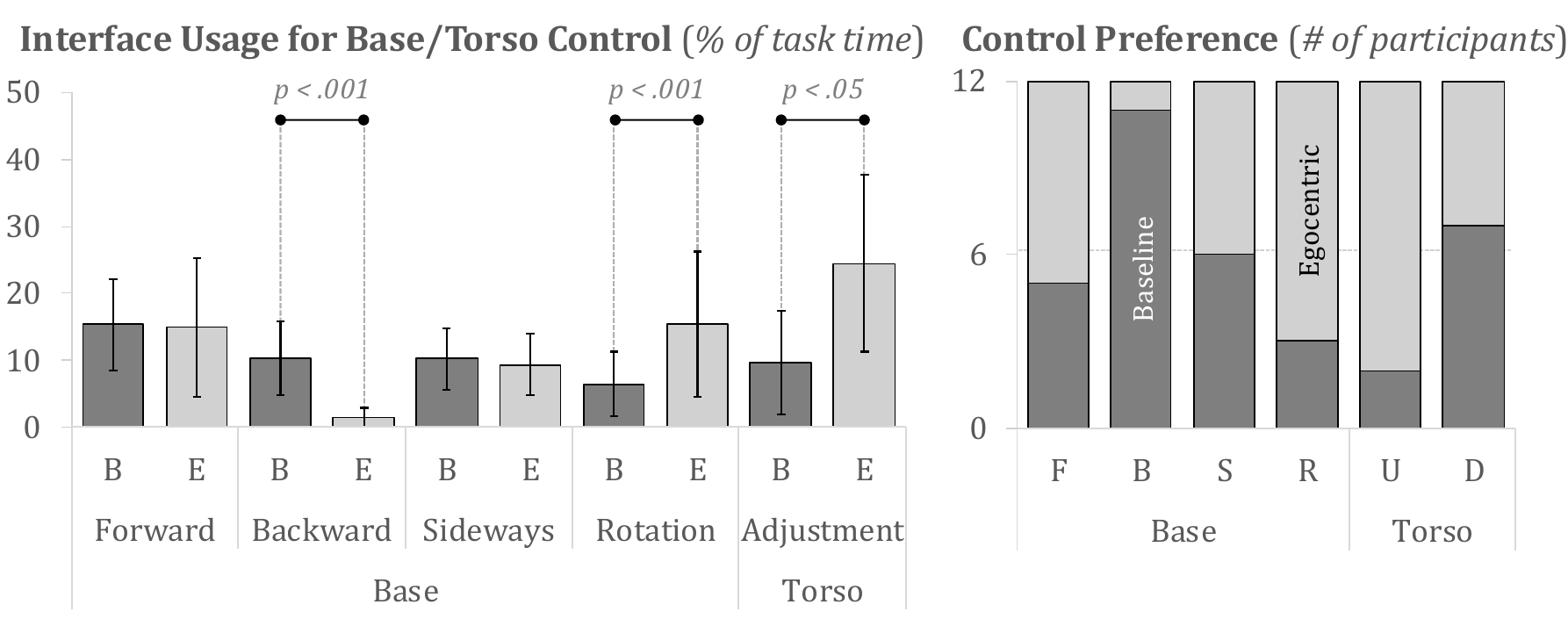}
    \caption{Baseline (B) and coupled egocentric (E) control usage and preference for moving the base forward (F), backward (B), sideways (S), rotating the base (R), and adjusting the torso up (U) and down (D) during the cleaning task.}
    \vspace{-3ex}
    \label{fig:phase2-distribution}
\end{figure} 

Correlation analyses showed that greater use of coupled egocentric control was associated with significantly shorter completion time for the complex cleaning task ($p < .001$) and lower overall NASA-TLX workload ($p < .05$), as shown in \fig{fig:phase2-regression}. These results indicate that participants who relied more on body-following control completed the task more efficiently and with less perceived workload.

Interface usage patterns further revealed how participants used the two control modes during the cleaning task. As shown in \fig{fig:phase2-distribution}, participants used coupled egocentric control significantly more often than the baseline for base rotation and torso adjustment. In contrast, there was no significant difference between the baseline and coupled egocentric interfaces for forward and sideways base movement. Notably, backward base movement was predominantly controlled using the baseline touchpad method ($p < .001$). Participants' stated preferences generally aligned with these usage patterns, with stronger preference for coupled egocentric control in torso adjustment, especially upward torso movement.

\textit{Discussion ---} Phase II demonstrates that coupled egocentric control can support complex, cluttered, home-care-inspired tasks by improving efficiency and reducing workload. The usage results also reveal an important distinction between perception-centered and manipulation-centered coupling. Participants strongly adopted perception-centered coupling, especially using head motion to control base rotation and torso adjustment, because these motions naturally extended the robot's viewpoint. Although torso motion could be driven by both head and arm movement, participants primarily used head-driven torso adjustment, suggesting that vertical body motion was often interpreted as part of active perception. 

Manipulation-centered coupling showed more mixed usage. Some participants preferred to manually position the robot near the workspace before manipulating objects, while others used coupled egocentric control to move the base and arms simultaneously for greater efficiency. The strongest exception was backward base motion, where participants preferred the baseline touchpad. This preference likely reflects the ergonomics of the input mapping: moving the arm backward can feel less natural and less efficient because it moves the hand away from the manipulation target. Overall, these findings suggest that body-following control is especially effective when the coupling aligns with natural perceptual or manipulation intent, while certain motions may still benefit from explicit manual commands.

\textit{Design Implications ---}
The Phase II results suggest that body-following control should remain interpretable, selectable, and easy to override in complex tasks. Participants strongly adopted head-driven torso adjustment and base rotation, indicating that perception-centered coupling can be enabled by default because it naturally extends the robot's viewpoint. In contrast, arm-driven base translation was more preference-dependent and sometimes better handled through explicit manual commands, especially for backward motion. These findings suggest that future whole-body teleoperation interfaces should preserve direct control over high-intent channels such as gaze and hands, while using simple, predictable, and interruptible coupling rules to automate supportive body motions.

%%%=============================================
\section{Conclusion}\label{sec:con}
%%%=============================================

In this paper, we presented \textit{coupled egocentric control} for whole-body robot teleoperation, a body-following approach in which the robot's torso and base follow the operator's gaze and hands. Instead of requiring users to explicitly command every torso and base adjustment, our system couples head motion with torso elevation and base rotation, and end-effector motion with torso adjustment and base translation. Through a user study with home-care-inspired tasks, we showed that coupled egocentric control improves object manipulation efficiency, reduces button-based control effort and arm singularities, lowers subjective workload, and increases ease of use, ease of learning, confidence, and preference for torso and base control. These findings suggest that allowing the robot body to follow the operator's perceptual and manipulation intent can make whole-body teleoperation more fluid and usable.

\textit{Limitations and Future Work ---} 
Although coupled egocentric control improved performance and usability over the baseline hybrid interface, several limitations remain. First, our current implementation relies on handheld VR controllers for arm input. A more natural interface may be controller-free, using hand tracking or gesture input through emerging mixed-reality devices. Future work will investigate how hand-tracking-based control can be combined with coupled egocentric control while preserving reliable manipulation and mode switching. Second, our study evaluated the approach on a mobile manipulator; future work should compare its effectiveness across different whole-body platforms, including humanoid robots and systems with legged mobility. Third, while our coupling rules were designed to be simple and predictable, future systems could adapt the coupling thresholds or control mappings based on task context, user preference, or robot state. Finally, coupled egocentric control could be integrated with ergonomic and workload-aware models, such as real-time physical workload estimation~\cite{lin2023impacts}, to reduce fatigue during prolonged teleoperation and provide more personalized body-following assistance.

% % % % % % % % % % % % % % % % % % % % % % % % % % % % % % % % % 
\bibliographystyle{IEEEtran}
\bibliography{references}

\end{document}